\documentclass{article}

\usepackage{microtype}
\usepackage{graphicx}
\usepackage{subcaption}
\usepackage{booktabs} 
\usepackage{multirow}
\usepackage{tabularx} 
\usepackage{enumitem}
\usepackage{hyperref}

\usepackage[accepted]{icml2026}

\usepackage{amsmath}
\usepackage{amssymb}
\usepackage{mathtools}
\usepackage{amsthm}
\definecolor{canyon}{RGB}{205, 92, 92}

\usepackage[capitalize,noabbrev]{cleveref}

\usepackage{caption}

\theoremstyle{plain}

\theoremstyle{definition}

\theoremstyle{remark}

\usepackage[textsize=tiny]{todonotes}

\icmltitlerunning{RiboSphere: Learning Unified and Efficient Representations of RNA Structures}

\begin{document}

\twocolumn[
  \icmltitle{RiboSphere: Learning Unified and Efficient Representations of RNA Structures}

  \icmlsetsymbol{equal}{*}
  \icmlsetsymbol{co}{\ensuremath{\dagger}}

  \begin{icmlauthorlist}
    \icmlauthor{Zhou Zhang}{equal,sch1}
    \icmlauthor{Hanqun Cao}{equal,sch2}
    \icmlauthor{Cheng Tan}{co,sch2}
    \icmlauthor{Fang Wu}{co,sch3}
    \icmlauthor{Pheng Ann Heng}{sch2}
    \icmlauthor{Tianfan Fu}{co,sch1}
  \end{icmlauthorlist}

  \icmlaffiliation{sch1}{State Key Laboratory for Novel Software Technology at Nanjing University, School of Computer Science, Nanjing University}
  \icmlaffiliation{sch2}{The Chinese University of Hong Kong}
  \icmlaffiliation{sch3}{Stanford University}

  \icmlcorrespondingauthor{Cheng Tan}{chengtan@cuhk.edu.hk}
  \icmlcorrespondingauthor{Fang Wu}{fangwu97@stanford.edu}
  \icmlcorrespondingauthor{Tianfan Fu}{futianfan@nju.edu.cn}

  \icmlkeywords{Machine Learning, ICML}

  \vskip 0.3in
]



\printAffiliationsAndNotice{
  \icmlEqualContribution\quad
  \textsuperscript{\ensuremath{\dagger}}Corresponding authors
}

\begin{abstract}
Accurate RNA structure modeling remains difficult because RNA backbones are highly flexible, non-canonical interactions are prevalent, and experimentally determined 3D structures are comparatively scarce. We introduce \emph{RiboSphere}, a framework that learns \emph{discrete} geometric representations of RNA by combining vector quantization with flow matching. Our design is motivated by the modular organization of RNA architecture: complex folds are composed from recurring structural motifs. RiboSphere uses a geometric transformer encoder trained using mean-centered coordinates and random rotation augmentation to produce geometry-aware features, which are discretized with finite scalar quantization (FSQ) into a finite vocabulary of latent codes. Conditioned on these discrete codes, a flow-matching decoder reconstructs atomic coordinates, enabling high-fidelity structure generation. We find that the learned code indices are enriched for specific RNA motifs, suggesting that the model captures motif-level compositional structure rather than acting as a purely compressive bottleneck. Across benchmarks, RiboSphere achieves strong performance in structure reconstruction (RMSD 1.25\,\AA, TM-score 0.84), and its pretrained discrete representations transfer effectively to inverse folding and RNA--ligand binding prediction, with robust generalization in data-scarce regimes. Code is available here: \url{https://github.com/Zhangz312/RiboSphere}
\end{abstract}

\section{Introduction}

Ribonucleic acid (RNA) plays multifaceted roles in living systems, ranging from genetic intermediaries to functional regulators \cite{goodall2021rna,morris2014rise}, with its diverse biological functions intimately linked to its complex three-dimensional conformations \cite{crick1970central}. Developing a universal RNA structural encoder capable of capturing intricate geometric constraints while maintaining robustness in data-scarce regimes has emerged as a central challenge at the intersection of computational biology and artificial intelligence \cite{townshend2020atom3d}. Despite the remarkable success of models such as AlphaFold in the protein domain \cite{jumper2021highly,abramson2024accurate}, RNA structure modeling continues to face formidable obstacles \cite{bernard2024has,martinovic2024comparative}.

\begin{figure}[t]
    \centering
    \includegraphics[width=\linewidth]{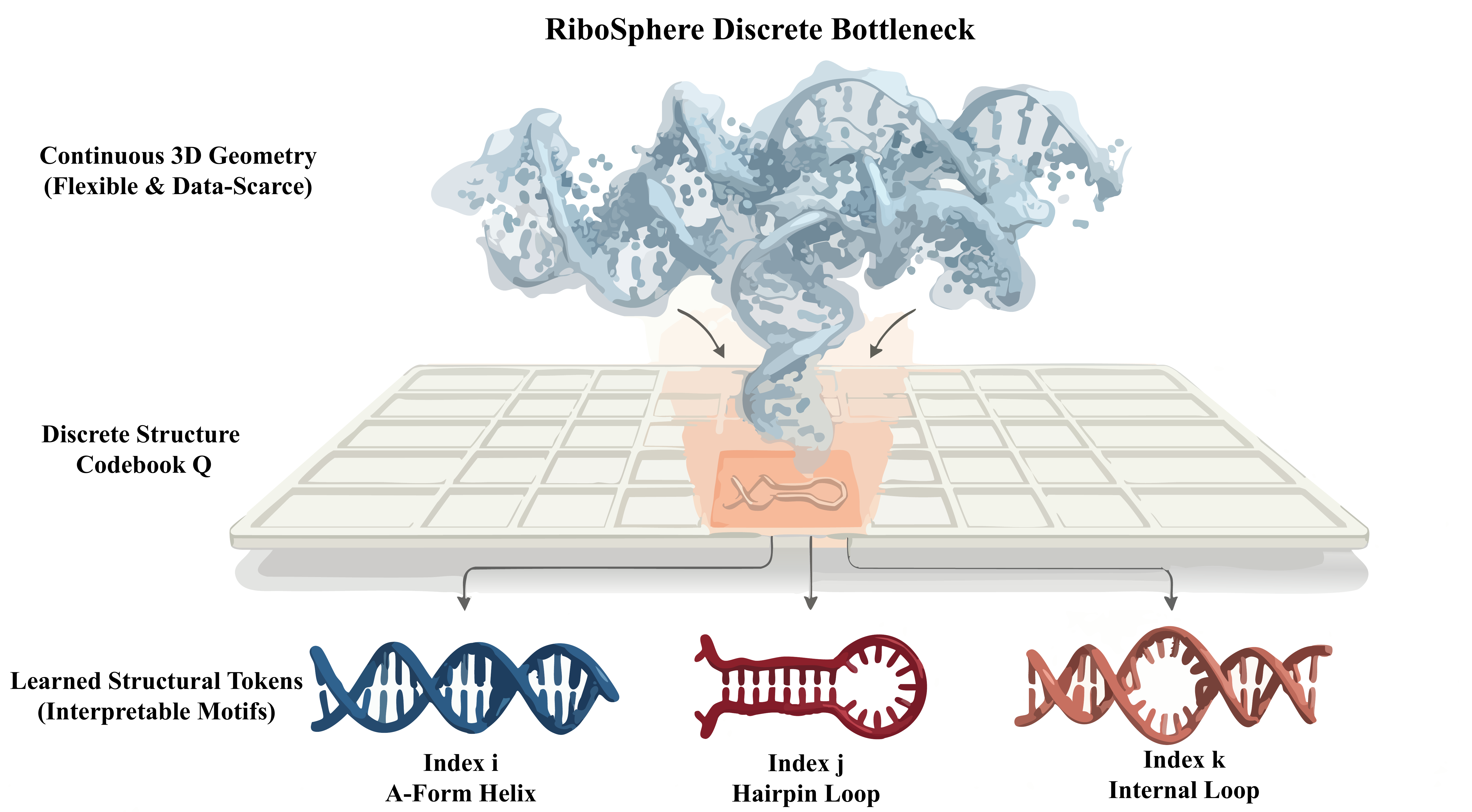} 
    \caption{\textbf{Overview of RiboSphere}: from continuous RNA geometry to discrete, interpretable structural units.}
    \label{fig:intro}
\end{figure}

Current methods for RNA structure representation face three key challenges:
\textbf{(1) Data scarcity.} While the Protein Data Bank contains over 200,000 protein structures, fewer than 6,000 RNA 3D structures have been experimentally resolved \cite{adamczyk2022rnasolo}. This order-of-magnitude gap renders conventional end-to-end representation learning prone to overfitting, and transfer learning from proteins offers limited remedy given fundamental differences in backbone chemistry.
\textbf{(2) Continuous space limitations.} Existing generative approaches operating in continuous latent spaces face challenges on sparse data manifolds \cite{ramakers2024novo}: VAEs suffer from posterior collapse with expressive decoders, while diffusion models may produce biophysically implausible conformations without strong geometric priors. The high backbone flexibility of RNA further exacerbates these issues.
\textbf{(3) Lack of interpretability.} Most learned representations function as black-box embeddings, lacking correspondence to biologically meaningful units. Yet RNA conformations are composed of recurring motifs---hairpin loops, internal loops, junctions---that serve as functional building blocks \cite{leontis2006building}. Capturing this modularity would improve both generalization and mechanistic understanding.

These challenges highlight a core bottleneck of existing RNA representations: continuous latent spaces struggle to generalize under data scarcity and fail to capture the modular, motif-based organization of RNA structures. As illustrated in Figure~\ref{fig:intro}, this motivates a shift from continuous RNA geometry toward discrete and interpretable structural units.

To address these challenges, we propose RiboSphere, a framework that learns discrete geometric representations of RNA by integrating Vector Quantization (VQ) \cite{van2017neural} with Flow Matching (FM) \cite{lipman2023flow}. Our key insight is that complex RNA conformations are not amorphous continua but rather compositional assemblies of a finite set of geometrically distinctive structural motifs, such as hairpin loops, internal loops, and pseudoknots \cite{leontis2006building,duarte2003rna}. RiboSphere leverages this modularity by projecting local geometries onto a learnable discrete codebook via Finite Scalar Quantization (FSQ), while a flow-based decoder reconstructs atomic coordinates from these discrete priors with sub-angstrom fidelity. 

Our main contributions are three-fold:
\begin{itemize}
    \item We propose RiboSphere, a VQ-Flow framework that combines a geometric transformer encoder with FSQ discretization and flow matching decoding. The FSQ bottleneck naturally avoids posterior collapse without auxiliary losses, and the asymmetric architecture (shallow encoder, deep decoder) encourages compression of key structural features into the discrete latent space.
    \item We demonstrate that the learned codebook captures biologically meaningful structure: token sequences exhibit high geometric consistency (RMSD $<$ 0.5\,\AA), and different RNA motifs (hairpin loops, internal loops, three-way junctions) show statistically distinct token distributions, suggesting the model learns interpretable structural primitives rather than arbitrary compression.
    \item Extensive experiments show that RiboSphere achieves state-of-the-art structure reconstruction (RMSD 1.25\,\AA, TM-score 0.84), the highest sequence recovery in inverse folding (63.0\%), and strong generalization in RNA-ligand binding prediction under out-of-distribution splits, outperforming GerNA-Bind by 2.6\% on the most challenging homology-fingerprint split.
\end{itemize}
\section{Related Work}
\subsection{RNA Structure Representation}
RNA 3D structure representation has evolved from computer vision-inspired volumetric grids to physics-aware geometric graphs. Early approaches leveraged 3D CNNs by discretizing molecular space into voxels to capture atomic densities  \cite{li2018rna3dcnn,peng2020bitenet,sun2024contrastive}. While these grid-based methods pioneered quality assessment and binding site detection, they suffer from sparsity and lack rotational invariance, requiring expensive data augmentation \cite{das2010atomic,li2019deepatom}. Alternative work explored backbone torsion angles to discretize folding into "structural alphabets" \cite{hershkovitz2003automated}, but these internal coordinates suffer from the "lever-arm" effect, where minor angular errors propagate into large global deviations.

The current frontier has shifted toward E(3)-equivariant Graph Neural Networks. By encoding physical symmetries directly into the architecture, models like ARES \cite{townshend2021geometric} achieve remarkable data efficiency, outperforming voxel-based methods even with minimal training samples. Recent frameworks including EquiRNA \cite{li2025size} and gRNAde \cite{joshi2025grnade} extend this paradigm through hierarchical representations and multi-state GNNs, enabling modeling of large-scale complexes and dynamic conformational ensembles. We build upon these geometric foundations with a discrete latent space that preserves symmetries while enabling efficient generative sampling.

\subsection{Autoencoders for Biomolecules}
Generative modeling for biomolecules has transitioned from continuous latent models to discrete tokenization frameworks that better align with the hierarchical nature of biological structures. Early research utilized VAEs with continuous Gaussian priors to capture evolutionary constraints \cite{riesselman2018deep}, but these models often suffer from posterior collapse when paired with expressive decoders and struggle to resolve the rugged energy structure landscapes \cite{ramakers2024novo}. VQ-VAEs \cite{xue2019supervised} emerged as a robust alternative, discretizing the latent space into a codebook of structural motifs. This paradigm has been successfully applied to protein engineering: FoldToken \cite{gao2025foldtoken} uses soft quantization to enable sub-Angstrom backbone reconstruction, while GCP-VQVAE \cite{pourmirzaei2025gcp} and ESM3 \cite{hayes2025simulating} integrate SE(3)-equivariant encoders to keep geometric consistency. Hybrid approaches like ProVQ \cite{liu2025designing} and SLM \cite{lu2025structure} further combine VQ-VAEs with diffusion models to capture dynamic conformational ensembles.

In the RNA domain, generative modeling is still navigating the balance between sequence-based fluency and structural fidelity. While GenerRNA \cite{zhao2024generrna} leverages large-scale language modeling to learn RNA "grammar" from sequences, structural models like Dfold \cite{ramakers2024novo} have only recently introduced VQ-VAEs for de novo 3D prediction. However, current RNA-specific models often rely on coarse-grained representations, such as three-class distance binning, which lack the resolution to model intricate non-canonical interactions. Furthermore, existing methods primarily focus on single-chain RNA folding, leaving a methodological gap in modeling multi-chain interfaces. While RiboSphere is currently trained and evaluated on single-chain RNA, the underlying discrete, geometry-complete quantization framework is conceptually compatible with multi-chain docking and interface modeling, which we plan to explore in future work.
\begin{figure*}[t]
    \centering
    \includegraphics[width=0.9\linewidth]{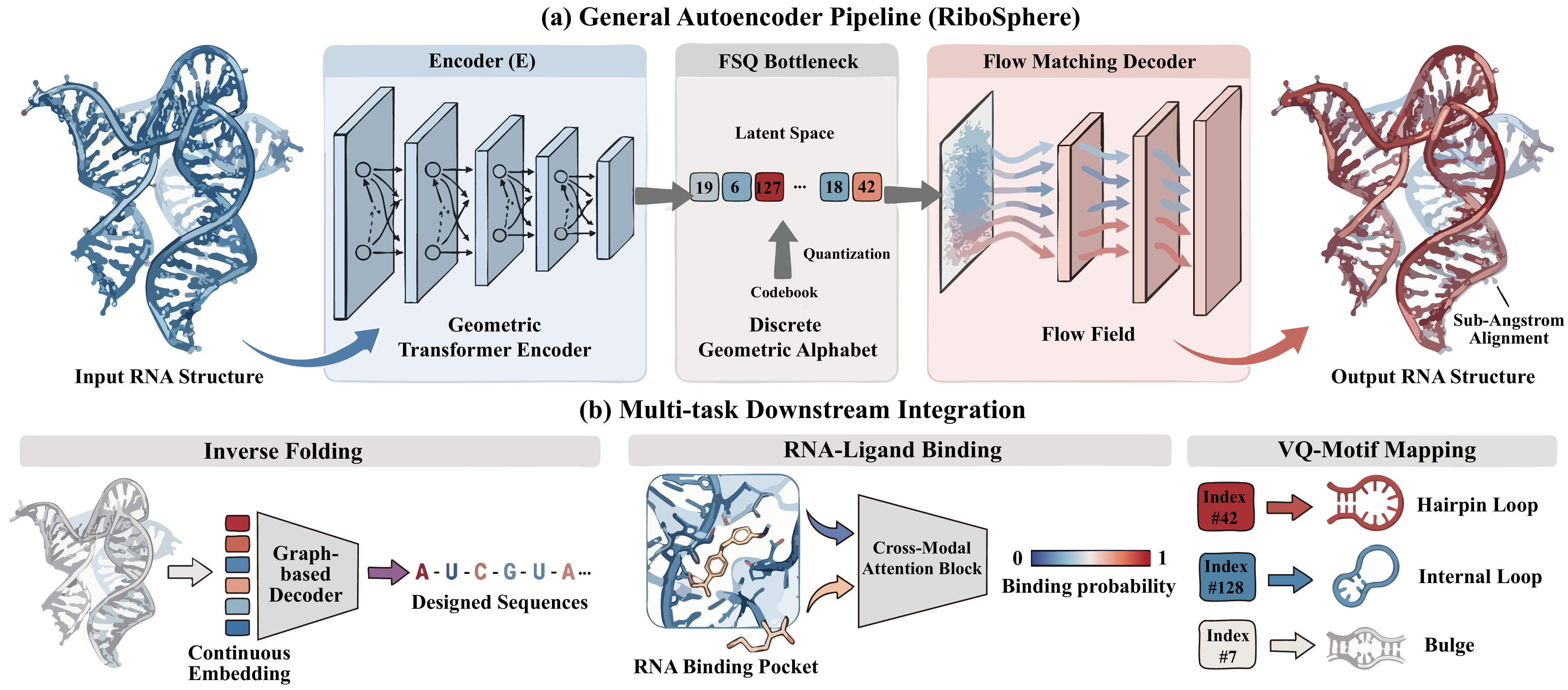} 
    \caption{Overall pipeline of \textbf{RiboSphere}. (a) \textbf{General autoencoder pipeline.} RNA atomic coordinates are encoded by a geometric transformer into continuous latent representations, which are discretized via Finite Scalar Quantization (FSQ) to obtain discrete geometric tokens. A flow-matching decoder reconstructs full 3D structures from the discrete representations, enabling high-fidelity structure reconstruction and serving as the pretraining objective. (b) \textbf{Multi-task downstream integration.} The pretrained encoder and quantizer are frozen and reused across downstream tasks. Discrete structural tokens and continuous embeddings are transferred to task-specific architectures for inverse folding and RNA--ligand binding prediction, providing a shared and interpretable geometric backbone.}
    \label{fig:framework}
\end{figure*}
\section{Method}
\subsection{Preliminaries}
We address the problem of learning representations and generating three-dimensional structures of RNA molecules. An RNA molecule is defined by its sequence $\mathbf{s} = (s_1, s_2, \ldots, s_L)$ where $s_i \in \{\text{A}, \text{U}, \text{C}, \text{G}\}$ denotes the nucleotide type, and its three-dimensional structure represented as atomic coordinates:
\begin{equation}
    \mathbf{x} \in \mathbb{R}^{L \times A \times 3},
\end{equation}
where $L$ is the sequence length and $A$ is the number of atoms per nucleotide. Depending on the modeling granularity, we define three atomic-level representation strategies:
\begin{itemize}
    \item \textbf{Single-atom model}: $A=1$, using only the C4' atom.
    \item \textbf{10-atom model}: $A=10$, with the backbone atom set $\{\text{P}, \text{C5}', \text{C4}', \text{C3}', \text{C2}', \text{C1}', \text{O5}', \text{O4}', \text{O3}', \text{O2}'\}$.
    \item \textbf{11-atom model}: $A=11$, extending the 10-atom backbone with a base-anchoring atom: N9 for purines or N1 for pyrimidines.
\end{itemize}

All input coordinates are mean-centered to remove global translation, and random rotations are applied during training to improve generalization across global orientations.

\paragraph{Learning Objectives.}
Our framework employs a two-stage learning paradigm. In the \textit{pretraining stage}, we learn a discrete geometric encoder via structure reconstruction, formulated as maximizing the likelihood of atomic coordinates given the discrete latent representation:
\begin{equation}
    \mathcal{L}_{\text{recon}} = -\mathbb{E}_{\mathbf{x} \sim p_{\text{data}}} \left[ \log p_\phi(\mathbf{x} \mid \hat{\mathbf{c}}) \right], \quad \hat{\mathbf{c}} = \mathcal{Q}(\mathcal{E}_\theta(\mathbf{x})),
\end{equation}
where $\mathcal{E}_\theta$ denotes the encoder, $\mathcal{Q}$ the vector quantization module, and $p_\phi$ the flow-based decoder distribution.

In the \textit{downstream tasks}, including inverse folding and RNA-ligand binding, we freeze the pretrained encoder and quantizer, transferring the learned discrete representations $\hat{\mathbf{c}}$ and continuous representations $\mathcal{E}_\theta(\mathbf{x})$ to task-specific objectives.

\subsection{Encoder Architecture and Featurization}

As illustrated in Figure~\ref{fig:framework}(a), the encoder $\mathcal{E}_\theta$ maps RNA atomic coordinates $\mathbf{x} \in \mathbb{R}^{L \times A \times 3}$ to a latent sequence $\mathbf{c} = (c_1, c_2, \ldots, c_L)$ with $c_i \in \mathbb{R}^{d}$:
\begin{equation}
    \mathbf{c} = \mathcal{E}_\theta(\mathbf{x}).
\end{equation}

We first apply mean-centering to obtain $\tilde{\mathbf{x}}$, then flatten the atomic coordinates of each nucleotide and project them to the encoder hidden dimension $d$:
\begin{equation}
    h_i = \text{MLP}_{\text{in}}(\text{vec}(\tilde{\mathbf{x}}_i)), \quad h_i \in \mathbb{R}^{d}.
\end{equation}

To capture inter-nucleotide spatial relationships and sequence topology, we construct pairwise features by combining discretized distance embeddings and relative positional encodings:
\begin{equation}
    p_{ij} = \text{MLP}_{\text{pair}}(\text{LN}(\mathbf{e}_{ij}^{\text{dist}} + \mathbf{e}_{ij}^{\text{pos}})),
\end{equation}
where $\mathbf{e}_{ij}^{\text{dist}}$ is obtained by bucketing pairwise Euclidean distances and embedding the resulting indices, and $\mathbf{e}_{ij}^{\text{pos}}$ encodes truncated relative sequence positions.

The encoder employs multi-head self-attention, injecting pairwise features $p_{ij}$ as geometric biases:
\begin{equation}
    \text{Att}(h_i, h_j) = \text{softmax}\left(\frac{(h_i \mathbf{W}_Q)(h_j \mathbf{W}_K)^\top}{\sqrt{d_k}} + p_{ij}\right) h_j \mathbf{W}_V,
\end{equation}
where $\mathbf{W}_Q, \mathbf{W}_K, \mathbf{W}_V$ are learnable projection matrices and $d_k$ is the attention head dimension. Through multi-layer stacking combined with a sliding window mechanism, the encoder effectively models both global and local spatial dependencies, yielding context-enriched representations:
\begin{equation}
    c_i = h_i + \sum_{j \neq i} \text{Att}(h_i, h_j), \quad i = 1, \ldots, L.
\end{equation}

\subsection{Vector Quantization Bottleneck}

The encoder output $\mathbf{c}$ is discretized via Finite Scalar Quantization (FSQ)~\cite{mentzer2023finite}:
\begin{equation}
    \hat{c} = \left\lfloor l/2 \right\rfloor \cdot \tanh(\text{Linear}(\mathbf{c})),
\end{equation}
where $l$ denotes the number of quantization levels per dimension. The quantized $\hat{\mathbf{c}}$ serves as a discrete geometric summary shared across all downstream tasks, with the straight-through estimator employed during training to enable gradient backpropagation.

\begin{table*}[t]
\centering
\caption{Comparison of RNA Structure Reconstruction and Quantization Methods.} 
\label{tab:results}
\resizebox{0.9\textwidth}{!}{
\begin{tabular}{l ccccccc cc}
\toprule
\multirow{2}{*}{\textbf{Method}} & \multirow{2}{*}{\textbf{\# Enc}} & \multirow{2}{*}{\textbf{\# Dec}} & \multirow{2}{*}{\textbf{Dim}} & \multirow{2}{*}{\textbf{\# Atoms}} & \multicolumn{3}{c}{\textbf{Structure}} & \multicolumn{2}{c}{\textbf{VQ}} \\
\cmidrule(lr){6-8} \cmidrule(lr){9-10}
& & & & & \textbf{RMSD} & \textbf{TM-score} & \textbf{lDDT} & \textbf{Codebook} & \textbf{\% Util.} \\
\midrule
E2-D8, D256, A1 & 2 & 8 & 256 & 1  & 2.14 & 0.70 & 0.73 & 240 & 100.0 \\
E2-D8, D256, A10 & 2 & 8 & 256 & 10 & 1.80 & 0.76 & 0.77 & 240 & 100.0 \\
E2-D8, D256, A11 & 2 & 8 & 256 & 11 & 2.05 & 0.71 & 0.73 & 240 & 100.0 \\
\midrule
E6-D6, D512, A1 & 6 & 6 & 512 & 1  & 1.88 & 0.73 & 0.75 & 1,000 & 95.0 \\
E6-D6, D512, A10 & 6 & 6 & 512 & 10 & 2.44 & 0.67 & 0.71 & 1,000 & 88.5 \\
E6-D6, D512, A11 & 6 & 6 & 512 & 11 & 2.71 & 0.68 & 0.72 & 1,000 & 98.9 \\
\midrule
E2-D8, D256, A1 & 2 & 8 & 256 & 1  & 1.88 & 0.75 & 0.76 & 1,000 & 80.7 \\
E2-D8, D256, A10 & 2 & 8 & 256 & 10 & 2.33 & 0.69 & 0.74 & 1,000 & 84.2 \\
E2-D8, D256, A11 & 2 & 8 & 256 & 11 & 1.60 & 0.78 & 0.79 & 1,000 & 88.0 \\
\midrule
E2-D8, D256, A1 & 2 & 8 & 256 & 1  & 1.25 & 0.84 & 0.83 & 4,375 & 39.8 \\
E2-D8, D256, A10 & 2 & 8 & 256 & 10 & 1.58 & 0.80 & 0.76 & 4,375 & 39.4 \\
E2-D8, D256, A11 & 2 & 8 & 256 & 11 & 1.35 & 0.82 & 0.82 & 4,375 & 39.9 \\
\midrule
E4-D8, D256, A11 & 4 & 8 & 256 & 11  & 2.26 & 0.69 & 0.73 & 240 & 100.0 \\
E4-D6, D256, A11 & 4 & 6 & 256 & 11 & 2.43 & 0.67 & 0.72 & 240 & 100.0 \\
E6-D4, D256, A11 & 6 & 4 & 256 & 11 & 2.93 & 0.60 & 0.67 & 240 & 100.0 \\
E8-D2, D256, A11 & 8 & 2 & 256 & 11 & 3.77 & 0.53 & 0.62 & 240 & 100.0 \\
\bottomrule
\end{tabular}%
}
\end{table*}     

\subsection{Decoder and Structural Reconstruction}

We adopt Flow Matching~\cite{lipman2023flow} for continuous generative modeling of 3D structures. Given randomly initialized noise conformations $\mathbf{x}_t \in \mathbb{R}^{L \times A \times 3}$ and discrete conditioning $\hat{\mathbf{c}}$, the decoder learns a time-dependent vector field:
\begin{equation}
    \mathcal{D}_\phi(\mathbf{x}_t, t, \hat{\mathbf{c}}) = v_\phi(\mathbf{x}_t, t, \hat{\mathbf{c}}), \quad t \in [0, 1].
\end{equation}

The flow matching formulation enables flexible incorporation of advanced sampling strategies at inference time. Euler integration yields the sampling trajectory:
\begin{align}
    \mathbf{x}_{t+\Delta t} &= \mathbf{x}_t + v_\phi(\mathbf{x}_t, t, \hat{\mathbf{c}}) \Delta t, \\
    & \text{where} \quad t \in \left\{0, \frac{1}{N}, \frac{2}{N}, \ldots, \frac{N-1}{N}\right\}. \nonumber
\end{align}

Classifier-free guidance~\cite{ho2022classifier} further enhances generation quality by amplifying sensitivity to conditioning information:
\begin{equation}
    \tilde{v}_\phi(\mathbf{x}_t, t, \hat{\mathbf{c}}) = v_\phi(\mathbf{x}_t, t, \hat{\mathbf{c}}) + g \cdot \left(v_\phi(\mathbf{x}_t, t, \hat{\mathbf{c}}) - v_\phi(\mathbf{x}_t, t, \varnothing)\right),
\end{equation}
where $g$ is the guidance weight and $\varnothing$ denotes the null conditioning vector.

For Gaussian flow, the vector field admits an analytical conversion to the score field:
\begin{equation}
    s_\phi(\mathbf{x}_t, t, \hat{\mathbf{c}}) = \frac{t \cdot v_\phi(\mathbf{x}_t, t, \hat{\mathbf{c}}) - \mathbf{x}_t}{1 - t},
\end{equation}
which enables stochastic differential equation (SDE) sampling for enhanced generation diversity:
\begin{equation}
    d\mathbf{x}_t = v_\phi(\mathbf{x}_t, t, \hat{\mathbf{c}}) dt + g(t) \eta \cdot s_\phi(\mathbf{x}_t, t, \hat{\mathbf{c}}) dt + \sqrt{2g(t)\gamma} \, d\mathbf{W}_t,
\end{equation}
where $\eta$ and $\gamma$ control the score gradient scaling and noise intensity, respectively.

The model is trained by minimizing the flow matching loss:
\begin{equation}
    \mathcal{L}_{\text{flow}} = \mathbb{E}_{\mathbf{x}_0, \mathbf{x}_1, t} \left[\left\| v_\phi(\mathbf{x}_t, t, \hat{\mathbf{c}}) - (\mathbf{x}_1 - \mathbf{x}_0) \right\|_2^2\right].
\end{equation}

\subsection{Inverse Folding}

We adapt discrete structural representations for the generation of sequences conditioned by structure, following the geometric inverse folding paradigm \cite{joshi2025grnade}.

\paragraph{Backbone Encoding.}
For inverse folding, we employ a 6-atom coarse-grained representation retaining the coordinates $\{\text{P}, \text{C5}', \text{C4}', \text{C3}', \text{O5}', \text{O3}'\}$ for each nucleotide. The frozen encoder maps the backbone coordinates to quantized structural features $\hat{\mathbf{c}} \in \mathbb{R}^{L \times d}$. These features provide a discrete geometric description learned with random rotation augmentation during pretraining.

\paragraph{Geometry Adapter.}
To supplement directional geometric relationships beyond discrete structural features, we introduce a Geometry Adapter that combines scalar-channel features with directional vector features. For each nucleotide $i$, we compute local coordinate frames from backbone atoms and extract directional vectors to form a joint scalar-vector representation:
\begin{equation}
    h_i^{\text{geo}} = (h_i^{\text{sca}}, \mathbf{v}_i), \quad h_i^{\text{sca}} \in \mathbb{R}^{d_s},
    \quad \mathbf{v}_i \in \mathbb{R}^{K \times 3}
\end{equation}
where $h_i^{\text{sca}}$ denotes scalar-channel features combining $\hat{c}_i$ with secondary structure and base-pairing priors, and $\mathbf{v}_i$ represents rotation-equivariant vector features derived from local frame orientations.

\paragraph{Autoregressive Decoding.}
The sequence decoder employs multi-layer Geometric Vector Perceptrons (GVP)~\cite{jing2021learning} for autoregressive prediction from the 5' to 3' end. At each position $i$, the decoder predicts:
\begin{equation}
    p_\psi(s_i \mid \hat{\mathbf{c}}, s_{<i}) = \text{softmax}\left(\text{GVP}_{\text{dec}}(h_i^{\text{geo}}, h_{<i}^{\text{seq}})\right)
\end{equation}
where $h_{<i}^{\text{seq}}$ encodes the partially decoded sequence context.

\paragraph{Training Objective.}
The model is optimized via cross-entropy loss with label smoothing ($\epsilon = 0.1$):
\begin{equation}
\begin{aligned}
    \mathcal{L}_{\text{inv}} = -\sum_{i=1}^{L} [& (1-\epsilon) \log p_\psi(s_i \mid \hat{\mathbf{c}}, s_{<i}) \\
    & + \frac{\epsilon}{4} \sum_{s' \in \mathcal{V}} \log p_\psi(s' \mid \hat{\mathbf{c}}, s_{<i})],
\end{aligned}
\end{equation}
where $\mathcal{V} = \{\text{A}, \text{U}, \text{C}, \text{G}\}$ is the nucleotide vocabulary.

\subsection{RNA-Ligand Binding Prediction}

We integrate the discrete structural representations into the GerNA-Bind framework~\cite{xia2025deciphering} for RNA-ligand binding prediction.

\paragraph{RNA Encoding.}
The RNA is represented through multimodal features. Our pretrained discrete structural embedding $\hat{\mathbf{c}}$ serves as the 3D geometric backbone, combined with:
\begin{equation}
    \mathbf{h}_{\text{RNA}} = \text{MLP}_{\text{fuse}}\left(\hat{\mathbf{c}} \| \mathbf{e}^{\text{1D}} \| \mathbf{e}^{\text{2D}}\right) \in \mathbb{R}^{L \times d}
\end{equation}
where $\mathbf{e}^{\text{1D}}$ denotes pretrained sequence embeddings and $\mathbf{e}^{\text{2D}}$ encodes secondary structure graph features.

\paragraph{Ligand Encoding.}
The ligand is encoded through both 2D molecular graph and 3D conformational information:
\begin{equation}
    \mathbf{h}_{\text{lig}} = \text{GraphTransformer}(\mathbf{G}_{\text{mol}}, \mathbf{X}_{\text{mol}}) \in \mathbb{R}^{N \times d}
\end{equation}
where $\mathbf{G}_{\text{mol}}$ is the molecular graph and $\mathbf{X}_{\text{mol}} \in \mathbb{R}^{N \times 3}$ contains 3D atomic coordinates.

\paragraph{Interaction Modeling.}
RNA-ligand interactions are modeled through a pairwise feature tensor refined by triangular attention:
\begin{align}
    \mathbf{Z}^{(0)}_{ij} &= \text{MLP}\left(\mathbf{h}_{\text{RNA},i} \| \mathbf{h}_{\text{lig},j} \| d_{ij}\right) \in \mathbb{R}^{d_p}, \\
    \mathbf{Z}^{(\ell+1)}_{ij} &= \mathbf{Z}^{(\ell)}_{ij} + \text{TriangleAttn}\left(\mathbf{Z}^{(\ell)}, \mathbf{h}_{\text{RNA}}, \mathbf{h}_{\text{lig}}\right),
\end{align}
where $d_{ij}$ encodes the spatial distance between RNA residue $i$ and ligand atom $j$.

\paragraph{Affinity Prediction.}
The final binding affinity is predicted via cross-modal attention pooling:
\begin{align}
    \mathbf{z}_{\text{global}} &= \text{CrossAttnPool}(\mathbf{Z}^{(L)}, \mathbf{h}_{\text{RNA}}, \mathbf{h}_{\text{lig}}), \\
    \hat{y} &= \sigma\left(\text{MLP}_{\text{pred}}(\mathbf{z}_{\text{global}})\right),
\end{align}
where $\sigma$ is the sigmoid function.

\paragraph{Training Objective.}
Following GerNA-Bind~\cite{xia2025deciphering}, we adopt an evidential deep learning objective:
\begin{equation}
\begin{aligned}
    \mathcal{L}_{\text{bind}} 
    = & \sum_{i} \left[(y_i - \hat{p}_i)^2 + \frac{\hat{p}_i(1-\hat{p}_i)}{S_i + 1}\right] \\
    & + \lambda_t \cdot \text{KL}\left[\text{Dir}(\tilde{\boldsymbol{\alpha}}_i) \| \text{Dir}(\mathbf{1})\right],
\end{aligned}
\end{equation}
where $\hat{p}_i$ and $S_i$ are derived from a Dirichlet-parameterized output, and $\lambda_t$ is an annealing coefficient.
\section{Experiments}

We evaluate RiboSphere across a series of tasks to assess the quality, generalization, and interpretability of its discrete RNA structural representations. Section~\ref{sec:setup} introduces the experimental settings and evaluation metrics. Section~\ref{sec:rna_reconstruct} studies the reconstruction of the RNA structure to analyze how the design of the codebook and the architectural choices affect the geometric expressivity. Section~\ref{sec:inverse_folding_experiment} evaluates the representations in RNA inverse folding from a generative perspective, while Section 4.4 examines their effectiveness in RNA–ligand binding prediction, particularly under out-of-distribution splits. Section 4.5 further analyzes the VQ codebook to validate the biological relevance and interpretability of the learned discrete structural tokens.

\subsection{Experimental Settings}
\label{sec:setup}

\noindent\textbf{Dataset. } 

\textbf{(i) Reconstruction}: The reconstruction task adopts the single-state split setting from gRNAde. Specifically, following the RNA structural clusters identified by Das et al. \cite{das2010atomic}, all RNA samples belonging to these clusters are collectively assigned to the test set, ensuring a strict structural decoupling between the test set and the training data. The remaining structural clusters that are not selected for the test set are randomly split into the training and validation sets. To further increase the effective sample size and enhance structural diversity, we explicitly expand multiple conformations associated with each RNA sequence, treating each conformation as an independent training instance. After this multi-conformation expansion, the resulting dataset consists of 11,183 training samples, 551 validation samples, and 239 test samples.

\textbf{(ii) Inverse folding}: The dataset follows gRNAde \cite{joshi2025grnade}, which is derived from RNASolo \cite{adamczyk2022rnasolo}, containing 4,223 unique RNA sequences and 12,011 structures. The dataset is split into single-state and multi-state clusters based on RNA structural flexibility.

\textbf{(iii) RNA-ligand binding}: The Robin dataset, derived from high-throughput microarray screening, focuses on 26 RNA targets and 1,893 drug-like molecules to provide 46,052 filtered interactions. Complementing this, the Biosensor dataset includes 191 synthetic ribozyme-aptamer RNAs and 89 drug-like molecules, totaling 16,999 interactions for evaluation. Rigorous evaluation is ensured through four split strategies: random, RNA homology-based, ligand fingerprint-based, and a combined homology-fingerprint approach \cite{xia2025deciphering}.

\noindent\textbf{Metrics. }
\textbf{Reconstruction}: The reconstruction task is primarily evaluated from the perspective of three-dimensional structural similarity. Performance is assessed using Root Mean Square Deviation (RMSD) to measure the average deviation between predicted and true atomic coordinates, TM-score to evaluate the similarity of the overall structural topology, and the local Distance Difference Test (lDDT) to quantify the accuracy of local geometric relationships in the predicted structures.

\textbf{Inverse folding}: 
Performance is evaluated using sequence metrics (recovery and diversity), 2D structural metrics from EternaFold (scMCC), and 3D structural metrics from RhoFold (RMSD, TM-score, pLDDT).

\textbf{RNA-ligand binding}: Binding specificity is measured by area under the receiver operating characteristic (AUROC).

\begin{table*}[ht]
\centering
\caption{Performance metrics for RNA sequence prediction
(\textbf{best}, \underline{second best}).}
\label{tab:sequence_metrics}
\resizebox{0.9\textwidth}{!}{
\begin{tabular}{l cccccc}
\toprule
\multirow{2}{*}{\textbf{Method}}
& \multicolumn{2}{c}{\textbf{Sequence}}
& \textbf{2D Struct.}
& \multicolumn{3}{c}{\textbf{3D Struct.}} \\
\cmidrule(lr){2-3}
\cmidrule(lr){4-4}
\cmidrule(lr){5-7}
& \textbf{Div.} ($\uparrow$)
& \textbf{Rec.} ($\uparrow$)
& \textbf{scMCC} ($\uparrow$)
& \textbf{RMSD} ($\downarrow$)
& \textbf{TM-score} ($\uparrow$)
& \textbf{pLDDT} ($\uparrow$) \\
\midrule
RiFold \cite{liu2025sentences}
& \textbf{0.89} & 0.416 & 0.28 & 17.06 & 0.12 & 0.50 \\

RDesign \cite{tan2024rdesign}
& 0.84 & 0.415 & 0.20 & 16.81 & 0.12 & 0.45 \\

RIDiffusion \cite{hou2025hyperbolic}
& \underline{0.85} & \underline{0.533} & 0.59
& \textbf{10.66} & \underline{0.25} & \underline{0.60} \\

gRNAde \cite{joshi2025grnade}
& 0.83 & 0.529 & \underline{0.60}
& \underline{11.45} & \textbf{0.29} & \textbf{0.62} \\
\midrule
RiboSphere
& 0.82 & \textbf{0.630} & \textbf{0.63}
& 11.54 & \textbf{0.29} & 0.55 \\
\bottomrule
\end{tabular}
}
\end{table*}

\begin{table*}[t]
\centering
\small 
\caption{Comparison of AUROC Metrics across Biosensor and Robin Datasets (\textbf{best}, \underline{second best}).}
\label{tab:auroc_comparison}
\begin{tabularx}{\textwidth}{l cccccccc}
\toprule
\multirow{3}{*}{\textbf{Method}} & \multicolumn{4}{c}{\textbf{AUROC-Biosensor}} & \multicolumn{4}{c}{\textbf{AUROC-Robin}} \\
\cmidrule(lr){2-5} \cmidrule(lr){6-9}
& \textbf{Random} & \textbf{RNA} & \textbf{Ligand} & \textbf{Homol. \&} & \textbf{Random} & \textbf{RNA} & \textbf{Ligand} & \textbf{Homol. \&} \\
& & \textbf{homology} & \textbf{fingerprint} & \textbf{fingerpr.} & & \textbf{homology} & \textbf{fingerprint} & \textbf{fingerpr.} \\
\midrule
\begin{tabular}{@{}l@{}}RSAPred\\\cite{krishnan2024reliable}\end{tabular} & 0.8764 & 0.7550 & 0.6707 & 0.6019 & 0.6327 & 0.5392 & 0.6320 & 0.4938 \\
\begin{tabular}{@{}l@{}}DeepDTIs\\\cite{wen2017deep}\end{tabular}     & 0.9300 & 0.8399 & 0.6813 & 0.6118 & 0.6302 & 0.5503 & 0.6290 & 0.4987 \\
\begin{tabular}{@{}l@{}}DeepConv-DTI\\\cite{lee2019deepconv}\end{tabular} & 0.9249 & 0.8427 & 0.6894 & 0.6213 & 0.6301 & 0.5625 & 0.6390 & 0.5104 \\
\begin{tabular}{@{}l@{}}GraphDTA\\\cite{nguyen2021graphdta}\end{tabular} & 0.8992 & 0.8284 & 0.7014 & 0.6370 & 0.6590 & 0.5528 & 0.6481 & 0.5510 \\
\begin{tabular}{@{}l@{}}GerNA-Bind\\\cite{xia2025deciphering}\end{tabular} & \textbf{0.9755} & \underline{0.9014} & \textbf{0.7723} & \underline{0.7279} & \textbf{0.7094} & \underline{0.6188} & \underline{0.6814} & \underline{0.6176} \\ \midrule 
RiboSphere   & \underline{0.9524} & \textbf{0.9031} & \underline{0.7406} & \textbf{0.7534} & \underline{0.6945} & \textbf{0.6335} & \textbf{0.6950} & \textbf{0.6349} \\
\bottomrule
\end{tabularx}
\end{table*}
\begin{figure*}[!ht]
    \centering
    \includegraphics[width=0.9\linewidth]{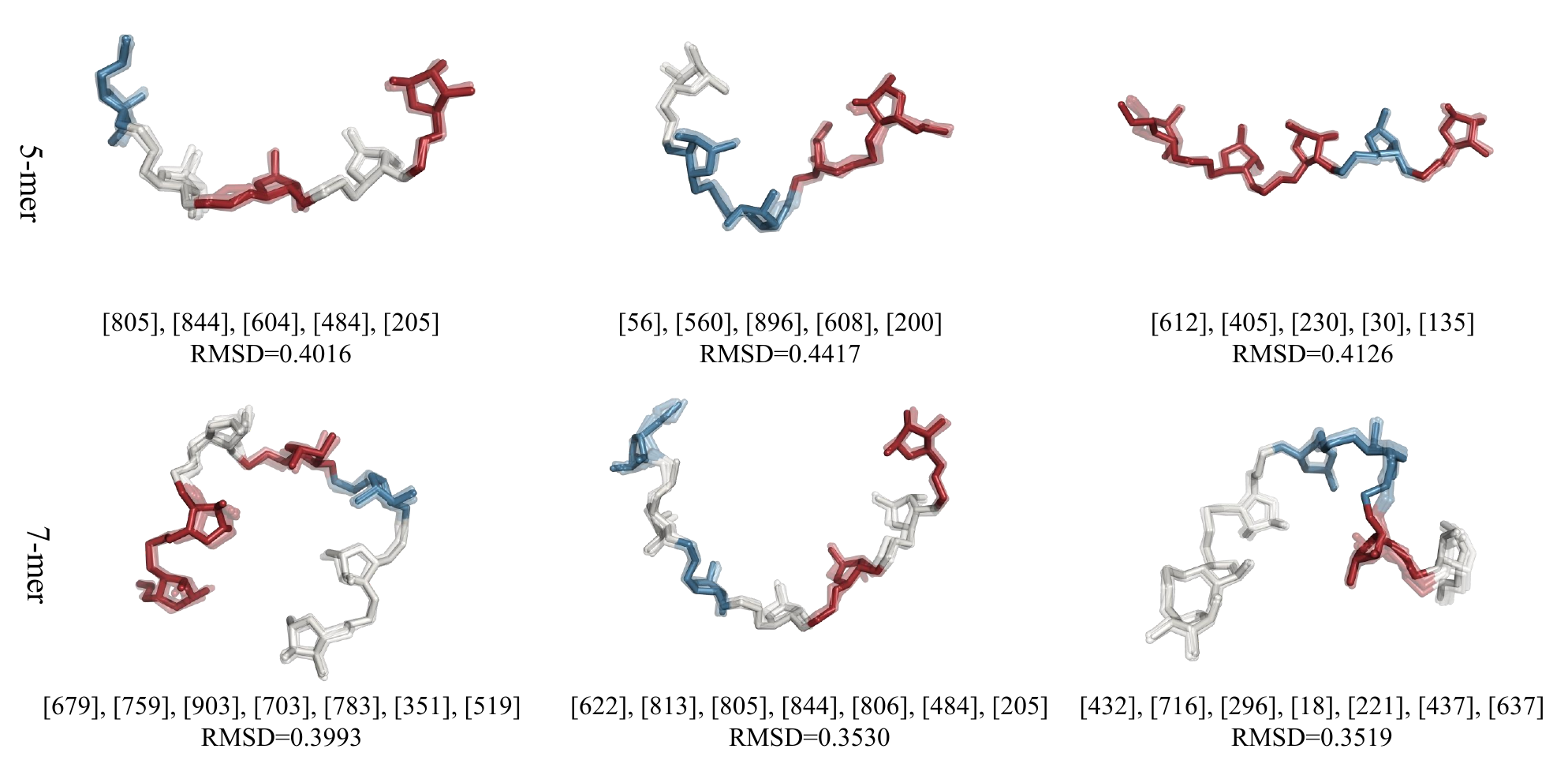} 
    \caption{Structural consistency of high-frequency VQ token sequences.}
    \label{fig: high-frequency}
\end{figure*}
\subsection{RNA Structure Reconstruction}
\label{sec:rna_reconstruct}
We evaluated RiboSphere under various configurations to identify the core design factors affecting reconstruction performance. As shown in Table~\ref{tab:results}, three key findings emerge:

\textbf{Atomic granularity matters, but requires sufficient codebook capacity.} Single-atom (C4$'$) representations preserve overall topology but lack base orientation information, introducing systematic errors at the atomic level. Adopting 10- or 11-atom representations enables the model to capture backbone-base configurations explicitly, yielding more accurate reconstruction. However, this benefit materializes only when the codebook is large enough to support the increased geometric diversity.

\textbf{Larger codebooks enable selective, semantically consistent encoding.} With a small codebook (240 tokens), the model achieves 100\% utilization but forces distinct conformations onto shared tokens, losing local geometric detail. As codebook size increases to 4,375, a sparse encoding pattern emerges spontaneously: utilization drops to $\sim$40\%, yet reconstruction improves. Rather than covering all tokens uniformly, RiboSphere selects a discriminative subset that captures semantically consistent local structures. This selective usage prevents posterior collapse while enhancing interpretability.

\textbf{Asymmetric architecture outperforms deeper encoders.} Contrary to intuition, performance does not increase monotonically with encoder depth. The best results arise from a shallow encoder (2 layers) paired with a deep decoder (8 layers). Overly expressive encoders can bypass the discrete bottleneck via complex continuous mappings, weakening the structural constraints imposed by quantization. A compact encoder forces key geometric features into the discrete latent space, improving representation robustness.

\subsection{Inverse Folding}
\label{sec:inverse_folding_experiment}
We apply RiboSphere's structural embeddings to RNA inverse folding. As shown in Table~\ref{tab:sequence_metrics}, RiboSphere achieves a sequence recovery rate of 63.0\%, substantially outperforming baselines including gRNAde (52.9\%) and RIDiffusion (53.3\%). Sequence diversity is slightly lower (0.82 vs. 0.89 for RiFold), reflecting a design choice: our model prioritizes structural fidelity over maximal diversity, which is often the primary objective in inverse folding.

For 2D self-consistency, designed sequences were folded with EternaFold. RiboSphere achieves the highest scMCC (0.633), indicating that its sequences best preserve secondary structure—consistent with the VQ bottleneck capturing local motifs. For 3D self-consistency, folded structures from RhoFold yield RMSD (11.54), TM-score (0.290), and pLDDT (0.548) comparable to gRNAde, while slightly lower pLDDT reflects predictor limitations. Overall, RiboSphere maintains strong sequence-structure consistency, particularly on structurally complex targets where the discrete codebook provides informative priors for non-canonical regions such as internal loops and junctions.

\begin{figure*}[t]
    \centering
    \includegraphics[width=0.9\linewidth]{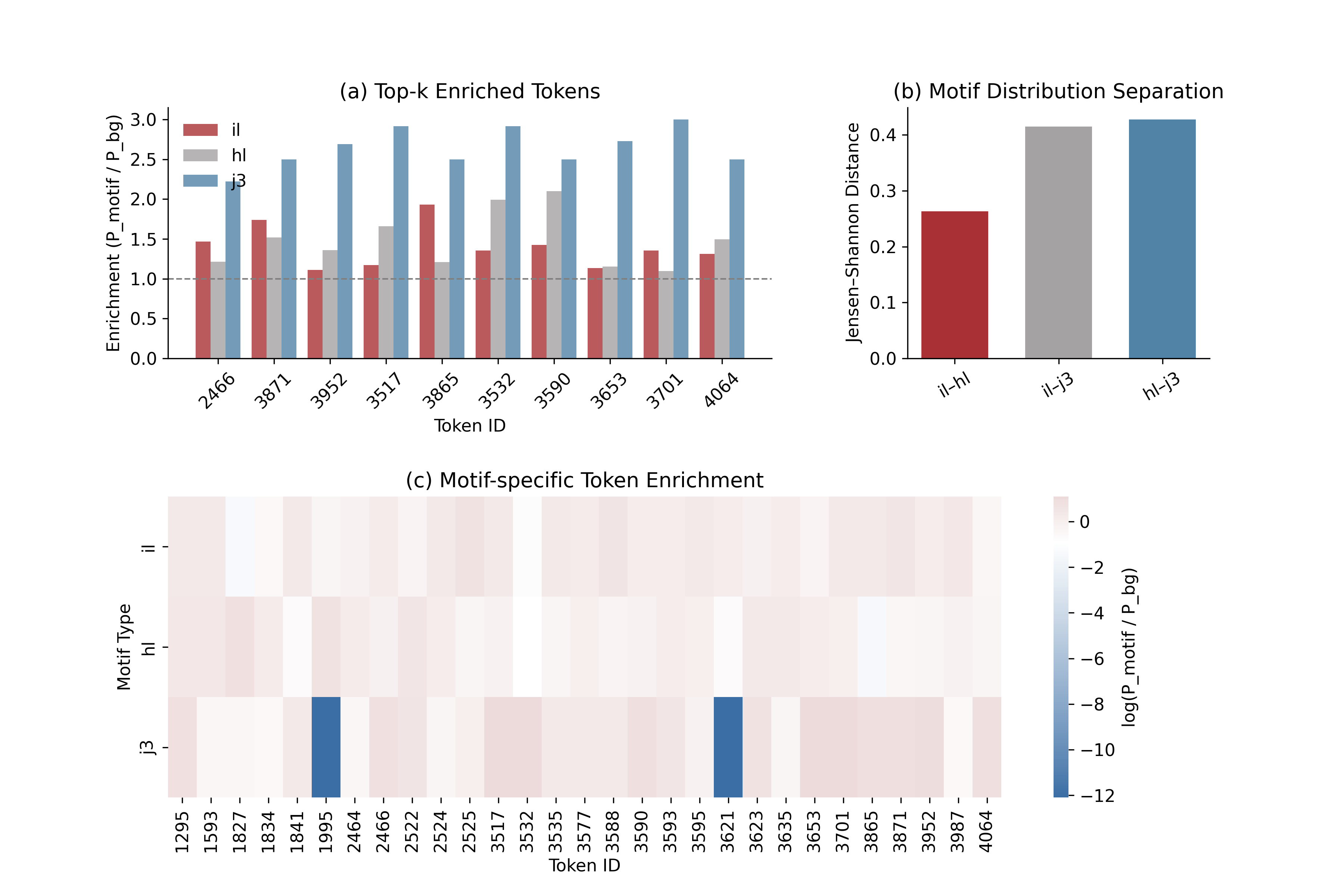} 
    \caption{Motif-Specific Token Distributions in the VQ Structural Space}
    \label{fig: token_distributions}
\end{figure*}

\subsection{RNA-Ligand Binding}

We evaluated RiboSphere on RNA-ligand binding prediction across two datasets (Biosensor and Robin) with four splitting strategies of increasing difficulty (Table~\ref{tab:auroc_comparison}). RiboSphere achieves the best performance on 5 of 8 tasks and ranks second on the remaining 3, consistently outperforming prior methods including GerNA-Bind.

Notably, RiboSphere demonstrates strong robustness under out-of-distribution (OOD) settings. On the most challenging Biosensor split (Homol.\ \& fingerpr.), RiboSphere achieves an AUROC of 0.7534, exceeding GerNA-Bind by 2.6\%. We attribute this advantage to the discrete bottleneck's denoising effect: codebook quantization filters conformational noise irrelevant to binding function, retaining only discriminative structural features such as binding-pocket geometries. This mechanism enables effective generalization to novel RNA folds unseen during training, suggesting that the learned discrete vocabulary captures transferable structural primitives rather than dataset-specific patterns.

\subsection{VQ-Structure Mapping}
We further analyzed the correspondence between discrete structural tokens obtained via VQ quantization and specific RNA motifs, aiming to answer a more fundamental question: do the discrete tokens learned by the model correspond to RNA local structural units with clear biophysical meaning? We validated this from two perspectives.

First, we examined the structural consistency of frequently occurring local VQ tokens. We first identified the most frequent 5-mer and 7-mer VQ token sequences across the dataset and traced all their geometric instances in the original RNA 3D structures. As shown in Figure \ref{fig: high-frequency}, structural segments mapped to the same discrete token sequences exhibit high spatial overlap, with average RMSD values remaining low (below 0.5 Å). At the structural level, some 5-mer tokens consistently correspond to specific types of backbone bends or torsional conformations, while certain 7-mer tokens capture more complete local structural units, such as the edges of hairpin loops or the stem–loop transition regions.

Second, starting from biologically defined RNA motifs, we analyzed their statistical signatures in the VQ token space. As illustrated in Figure~\ref{fig: token_distributions}, different motif types show clear deviations from the background token distribution. Internal loops (IL) and hairpin loops (HL) exhibit moderate KL divergence values of 0.11 and 0.12, respectively, indicating that their token usage is partially constrained while still retaining substantial geometric variability. In contrast, three-way junctions (J3) display a much higher KL divergence (0.70) and strong enrichment on a small subset of tokens, reflecting their highly constrained spatial configurations.

Furthermore, tokens across different motif types are well separated in terms of Jensen–Shannon distance, with the distances between J3 and other motifs being significantly larger than that between IL and HL. These results suggest that, although individual motif instances do not map to identical discrete token sequences, VQ tokens capture motif-specific local structural characteristics at a statistical level.

\section{Conclusion}
In this work, we introduced RiboSphere, a generative framework based on flow matching designed specifically for the reconstruction of complex RNA structures. Our empirical evaluations demonstrate that RiboSphere achieves state-of-the-art performance in structure modeling. Beyond reconstruction, we established that RiboSphere serves as a robust foundation for broader structural understanding tasks, including inverse folding and RNA-ligand binding affinity prediction.
Moving forward, we aim to extend our structural representations to accommodate a wider array of downstream tasks, such as zero-shot function annotation and the modeling of dynamic RNA-protein complexes.

\section*{Acknowledgements}
Zhou Zhang and Tianfan Fu are supported by Young Scientists Fund (C Class) of the National Natural Science Foundation of China (Grant No. 62506154), the Fundamental Research Funds for the Central Universities and Nanjing University International Collaboration Initiative (Grant No. 020214380129) and the “111 Center” (No. B26023). The work described in this paper was also partially supported by the Research Grants Council of the Hong Kong Special Administrative Region, China, under Project T45-401/22-N.

\section*{Impact Statement}
This work proposes a discrete geometric representation for RNA structure modeling, which improves the ability to model RNA structures and functions under data-scarce conditions and has the potential to advance research in RNA design, drug discovery, and synthetic biology. By learning a biophysically interpretable structural codebook, the method offers a new perspective for modular modeling of complex biological macromolecules. However, without experimental validation, the model’s predictions may be biologically implausible or misleading if over-interpreted. Overall, this work is expected to contribute positively to methodological advances in computational biology, without posing direct or significant societal risks.


\bibliography{example_paper}

@article{joshi2025grnade,
  title={grnade: Geometric deep learning for 3d rna inverse design},
  author={Joshi, Chaitanya K and Jamasb, Arian R and Vi{\~n}as, Ramon and Harris, Charles and Mathis, Simon V and Morehead, Alex and Anand, Rishabh and Li{\`o}, Pietro},
  journal={bioRxiv},
  pages={2024--03},
  year={2025}
}

@article{adamczyk2022rnasolo,
  title={RNAsolo: a repository of cleaned PDB-derived RNA 3D structures},
  author={Adamczyk, Bartosz and Antczak, Maciej and Szachniuk, Marta},
  journal={Bioinformatics},
  volume={38},
  number={14},
  pages={3668--3670},
  year={2022},
  publisher={Oxford University Press}
}

@article{das2010atomic,
  title={Atomic accuracy in predicting and designing noncanonical RNA structure},
  author={Das, Rhiju and Karanicolas, John and Baker, David},
  journal={Nature methods},
  volume={7},
  number={4},
  pages={291--294},
  year={2010},
  publisher={Nature Publishing Group US New York}
}

@inproceedings{li2025size,
  title={Size-generalizable RNA structure evaluation by exploring hierarchical geometries},
  author={Li, Zongzhao and Cen, Jiacheng and Huang, Wenbing and Wang, Taifeng and Song, Le},
  booktitle={The Thirteenth International Conference on Learning Representations},
  year={2025}
}

@article{li2018rna3dcnn,
  title={RNA3DCNN: Local and global quality assessments of RNA 3D structures using 3D deep convolutional neural networks},
  author={Li, Jun and Zhu, Wei and Wang, Jun and Li, Wenfei and Gong, Sheng and Zhang, Jian and Wang, Wei},
  journal={PLoS computational biology},
  volume={14},
  number={11},
  pages={e1006514},
  year={2018},
  publisher={Public Library of Science San Francisco, CA USA}
}

@inproceedings{peng2020bitenet,
  title={BiteNet: bidirectional temporal encoder network to predict medical outcomes},
  author={Peng, Xueping and Long, Guodong and Shen, Tao and Wang, Sen and Jiang, Jing and Zhang, Chengqi},
  booktitle={2020 IEEE International Conference on Data Mining (ICDM)},
  pages={412--421},
  year={2020},
  organization={IEEE}
}

@article{sun2024contrastive,
  title={Contrastive pre-training and 3D convolution neural network for RNA and small molecule binding affinity prediction},
  author={Sun, Saisai and Gao, Lin},
  journal={Bioinformatics},
  volume={40},
  number={4},
  pages={btae155},
  year={2024},
  publisher={Oxford University Press}
}

@inproceedings{li2019deepatom,
  title={DeepAtom: a framework for protein-ligand binding affinity prediction},
  author={Li, Yanjun and Rezaei, Mohammad A and Li, Chenglong and Li, Xiaolin},
  booktitle={2019 IEEE international conference on bioinformatics and biomedicine (BIBM)},
  pages={303--310},
  year={2019},
  organization={IEEE}
}

@article{hershkovitz2003automated,
  title={Automated identification of RNA conformational motifs: theory and application to the HM LSU 23S rRNA},
  author={Hershkovitz, Eli and Tannenbaum, Emmanuel and Howerton, Shelley B and Sheth, Ajay and Tannenbaum, Allen and Williams, Loren Dean},
  journal={Nucleic Acids Research},
  volume={31},
  number={21},
  pages={6249--6257},
  year={2003},
  publisher={Oxford University Press}
}

@article{townshend2021geometric,
  title={Geometric deep learning of RNA structure},
  author={Townshend, Raphael JL and Eismann, Stephan and Watkins, Andrew M and Rangan, Ramya and Karelina, Masha and Das, Rhiju and Dror, Ron O},
  journal={Science},
  volume={373},
  number={6558},
  pages={1047--1051},
  year={2021},
  publisher={American Association for the Advancement of Science}
}

@article{ramakers2024novo,
  title={De novo prediction of RNA 3D structures with deep generative models},
  author={Ramakers, Julius and Blum, Christopher Frederik and K{\"o}nig, Sabrina and Harmeling, Stefan and Kollmann, Markus},
  journal={Plos one},
  volume={19},
  number={2},
  pages={e0297105},
  year={2024},
  publisher={Public Library of Science San Francisco, CA USA}
}

@article{riesselman2018deep,
  title={Deep generative models of genetic variation capture the effects of mutations},
  author={Riesselman, Adam J and Ingraham, John B and Marks, Debora S},
  journal={Nature methods},
  volume={15},
  number={10},
  pages={816--822},
  year={2018},
  publisher={Nature Publishing Group US New York}
}

@article{xue2019supervised,
  title={Supervised vector quantized variational autoencoder for learning interpretable global representations},
  author={Xue, Yifan and Ding, Michael and Lu, Xinghua},
  journal={arXiv preprint arXiv:1909.11124},
  year={2019}
}

@inproceedings{gao2025foldtoken,
  title={Foldtoken: Learning protein language via vector quantization and beyond},
  author={Gao, Zhangyang and Tan, Cheng and Wang, Jue and Huang, Yufei and Wu, Lirong and Li, Stan Z},
  booktitle={Proceedings of the AAAI Conference on Artificial Intelligence},
  volume={39},
  number={1},
  pages={219--227},
  year={2025}
}

@article{pourmirzaei2025gcp,
  title={GCP-VQVAE: A Geometry-Complete Language for Protein 3D Structure},
  author={Pourmirzaei, Mahdi and Morehead, Alex and Esmaili, Farzaneh and Ren, Jarett and Pourmirzaei, Mohammadreza and Xu, Dong},
  journal={bioRxiv},
  pages={2025--10},
  year={2025},
  publisher={Cold Spring Harbor Laboratory}
}

@article{liu2025designing,
  title={Designing flexible protein structures and sampling protein conformations with a unified model using vector quantization and diffusion},
  author={Liu, Yufeng and Chen, Linghui and Chen, Quan and Liu, Haiyan},
  journal={National Science Review},
  volume={12},
  number={11},
  pages={nwaf290},
  year={2025},
  publisher={Oxford University Press}
}

@article{zhao2024generrna,
  title={GenerRNA: A generative pre-trained language model for de novo RNA design},
  author={Zhao, Yichong and Oono, Kenta and Takizawa, Hiroki and Kotera, Masaaki},
  journal={PLoS One},
  volume={19},
  number={10},
  pages={e0310814},
  year={2024},
  publisher={Public Library of Science San Francisco, CA USA}
}

@article{hayes2025simulating,
  title={Simulating 500 million years of evolution with a language model},
  author={Hayes, Thomas and Rao, Roshan and Akin, Halil and Sofroniew, Nicholas J and Oktay, Deniz and Lin, Zeming and Verkuil, Robert and Tran, Vincent Q and Deaton, Jonathan and Wiggert, Marius and others},
  journal={Science},
  volume={387},
  number={6736},
  pages={850--858},
  year={2025},
  publisher={American Association for the Advancement of Science}
}

@inproceedings{
lu2025structure,
title={Structure Language Models for Protein Conformation Generation},
author={Jiarui Lu and Xiaoyin Chen and Stephen Zhewen Lu and Chence Shi and Hongyu Guo and Yoshua Bengio and Jian Tang},
booktitle={The Thirteenth International Conference on Learning Representations},
year={2025},
url={https://openreview.net/forum?id=OzUNDnpQyd}
}

@article{crick1970central,
  title={Central dogma of molecular biology},
  author={Crick, Francis},
  journal={Nature},
  volume={227},
  number={5258},
  pages={561--563},
  year={1970},
  publisher={Nature Publishing Group UK London}
}

@article{townshend2020atom3d,
  title={Atom3d: Tasks on molecules in three dimensions},
  author={Townshend, Raphael JL and V{\"o}gele, Martin and Suriana, Patricia and Derry, Alexander and Powers, Alexander and Laloudakis, Yianni and Balachandar, Sidhika and Jing, Bowen and Anderson, Brandon and Eismann, Stephan and others},
  journal={arXiv preprint arXiv:2012.04035},
  year={2020}
}

@article{abramson2024accurate,
  title={Accurate structure prediction of biomolecular interactions with AlphaFold 3},
  author={Abramson, Josh and Adler, Jonas and Dunger, Jack and Evans, Richard and Green, Tim and Pritzel, Alexander and Ronneberger, Olaf and Willmore, Lindsay and Ballard, Andrew J and Bambrick, Joshua and others},
  journal={Nature},
  volume={630},
  number={8016},
  pages={493--500},
  year={2024},
  publisher={Nature Publishing Group UK London}
}

@article{jumper2021highly,
  title={Highly accurate protein structure prediction with AlphaFold},
  author={Jumper, John and Evans, Richard and Pritzel, Alexander and Green, Tim and Figurnov, Michael and Ronneberger, Olaf and Tunyasuvunakool, Kathryn and Bates, Russ and {\v{Z}}{\'\i}dek, Augustin and Potapenko, Anna and others},
  journal={nature},
  volume={596},
  number={7873},
  pages={583--589},
  year={2021},
  publisher={Nature Publishing Group UK London}
}

@article{bernard2024has,
  title={Has AlphaFold 3 reached its success for RNAs?},
  author={Bernard, Cl{\'e}ment and Postic, Guillaume and Ghannay, Sahar and Tahi, Fariza},
  journal={bioRxiv},
  pages={2024--06},
  year={2024},
  publisher={Cold Spring Harbor Laboratory}
}

@article{leontis2006building,
  title={The building blocks and motifs of RNA architecture},
  author={Leontis, Neocles B and Lescoute, Aurelie and Westhof, Eric},
  journal={Current opinion in structural biology},
  volume={16},
  number={3},
  pages={279--287},
  year={2006},
  publisher={Elsevier}
}

@article{duarte2003rna,
  title={RNA structure comparison, motif search and discovery using a reduced representation of RNA conformational space},
  author={Duarte, Carlos M and Wadley, Leven M and Pyle, Anna Marie},
  journal={Nucleic Acids Research},
  volume={31},
  number={16},
  pages={4755--4761},
  year={2003},
  publisher={Oxford University Press}
}

@article{van2017neural,
  title={Neural discrete representation learning},
  author={Van Den Oord, Aaron and Vinyals, Oriol and others},
  journal={Advances in neural information processing systems},
  volume={30},
  year={2017}
}

@inproceedings{
lipman2023flow,
title={Flow Matching for Generative Modeling},
author={Yaron Lipman and Ricky T. Q. Chen and Heli Ben-Hamu and Maximilian Nickel and Matthew Le},
booktitle={The Eleventh International Conference on Learning Representations },
year={2023},
url={https://openreview.net/forum?id=PqvMRDCJT9t}
}

@article{goodall2021rna,
  title={RNA in cancer},
  author={Goodall, Gregory J and Wickramasinghe, Vihandha O},
  journal={Nature Reviews Cancer},
  volume={21},
  number={1},
  pages={22--36},
  year={2021},
  publisher={Nature Publishing Group UK London}
}

@article{morris2014rise,
  title={The rise of regulatory RNA},
  author={Morris, Kevin V and Mattick, John S},
  journal={Nature Reviews Genetics},
  volume={15},
  number={6},
  pages={423--437},
  year={2014},
  publisher={Nature Publishing Group UK London}
}

@article{martinovic2024comparative,
  title={A Comparative Review of Deep Learning Methods for RNA Tertiary Structure Prediction},
  author={Martinovi{\'c}, Ivona and Vla{\v{s}}i{\'c}, Tin and Li, Yang and Hooi, Bryan and Zhang, Yang and {\v{S}}iki{\'c}, Mile},
  journal={bioRxiv},
  pages={2024--11},
  year={2024},
  publisher={Cold Spring Harbor Laboratory}
}

@article{mentzer2023finite,
  title={Finite scalar quantization: Vq-vae made simple},
  author={Mentzer, Fabian and Minnen, David and Agustsson, Eirikur and Tschannen, Michael},
  journal={arXiv preprint arXiv:2309.15505},
  year={2023}
}

@article{ho2022classifier,
  title={Classifier-free diffusion guidance},
  author={Ho, Jonathan and Salimans, Tim},
  journal={arXiv preprint arXiv:2207.12598},
  year={2022}
}

@inproceedings{
jing2021learning,
title={Learning from Protein Structure with Geometric Vector Perceptrons},
author={Bowen Jing and Stephan Eismann and Patricia Suriana and Raphael John Lamarre Townshend and Ron Dror},
booktitle={International Conference on Learning Representations},
year={2021},
url={https://openreview.net/forum?id=1YLJDvSx6J4}
}

@article{dauparas2022robust,
  title={Robust deep learning--based protein sequence design using ProteinMPNN},
  author={Dauparas, Justas and Anishchenko, Ivan and Bennett, Nathaniel and Bai, Hua and Ragotte, Robert J and Milles, Lukas F and Wicky, Basile IM and Courbet, Alexis and de Haas, Rob J and Bethel, Neville and others},
  journal={Science},
  volume={378},
  number={6615},
  pages={49--56},
  year={2022},
  publisher={American Association for the Advancement of Science}
}

@article{zhang2004scoring,
  title={Scoring function for automated assessment of protein structure template quality},
  author={Zhang, Yang and Skolnick, Jeffrey},
  journal={Proteins: Structure, Function, and Bioinformatics},
  volume={57},
  number={4},
  pages={702--710},
  year={2004},
  publisher={Wiley Online Library}
}

@article{liu2025sentences,
  title={From sentences to sequences: Rethinking languages in biological system},
  author={Liu, Ke and Shen, Shuaike and Chen, Hao},
  journal={arXiv preprint arXiv:2507.00953},
  year={2025}
}

@inproceedings{tan2024rdesign,
  title={RDesign: Hierarchical data-efficient representation learning for tertiary structure-based RNA design},
  author={Tan, Cheng and Zhang, Yijie and Gao, Zhangyang and Hu, Bozhen and Li, Siyuan and Liu, Zicheng and Li, Stan Z},
  booktitle={International Conference on Learning Representations},
  volume={2024},
  pages={10304--10333},
  year={2024}
}

@article{hou2025hyperbolic,
  title={A hyperbolic discrete diffusion 3D rna inverse folding model for functional RNA design},
  author={Hou, Dongyue and Zhang, Shuai and Ma, Mengyao and Lin, Hanbo and Wan, Zheng and Zhao, Hui and Zhou, Ruian and He, Xiao and Wei, Xian and Ju, Dianwen and others},
  journal={Journal of Chemical Information and Modeling},
  volume={65},
  number={13},
  pages={6568--6584},
  year={2025},
  publisher={ACS Publications}
}

@article{krishnan2024reliable,
  title={Reliable method for predicting the binding affinity of RNA-small molecule interactions using machine learning},
  author={Krishnan, Sowmya R and Roy, Arijit and Gromiha, M Michael},
  journal={Briefings in Bioinformatics},
  volume={25},
  number={2},
  pages={bbae002},
  year={2024},
  publisher={Oxford University Press}
}

@article{wen2017deep,
  title={Deep-learning-based drug--target interaction prediction},
  author={Wen, Ming and Zhang, Zhimin and Niu, Shaoyu and Sha, Haozhi and Yang, Ruihan and Yun, Yonghuan and Lu, Hongmei},
  journal={Journal of proteome research},
  volume={16},
  number={4},
  pages={1401--1409},
  year={2017},
  publisher={ACS Publications}
}

@article{lee2019deepconv,
  title={DeepConv-DTI: Prediction of drug-target interactions via deep learning with convolution on protein sequences},
  author={Lee, Ingoo and Keum, Jongsoo and Nam, Hojung},
  journal={PLoS computational biology},
  volume={15},
  number={6},
  pages={e1007129},
  year={2019},
  publisher={Public Library of Science San Francisco, CA USA}
}

@article{nguyen2021graphdta,
  title={GraphDTA: predicting drug--target binding affinity with graph neural networks},
  author={Nguyen, Thin and Le, Hang and Quinn, Thomas P and Nguyen, Tri and Le, Thuc Duy and Venkatesh, Svetha},
  journal={Bioinformatics},
  volume={37},
  number={8},
  pages={1140--1147},
  year={2021},
  publisher={Oxford University Press}
}

@article{xia2025deciphering,
  title={Deciphering RNA--ligand binding specificity with GerNA-Bind},
  author={Xia, Yunpeng and Li, Jiayi and Chu, Yi-Ting and Rao, Jiahua and Chen, Jing and Hua, Chenqing and Yu, Dong-Jun and Chen, Xiu-Cai and Zheng, Shuangjia},
  journal={Nature Machine Intelligence},
  pages={1--13},
  year={2025},
  publisher={Nature Publishing Group UK London}
}
\bibliographystyle{icml2026}

\newpage
\appendix
\onecolumn
\section{Metric}
This section describes all metrics used to assess model performance across the different tasks studied in this paper, together with their formal definitions and computation procedures.

\subsection{Reconstruction Metrics}
For the structure reconstruction task, model performance is primarily evaluated from the perspective of three-dimensional geometric similarity. We report the following complementary metrics.

\paragraph{Root Mean Square Deviation (RMSD).} 
RMSD measures the average deviation between predicted and ground-truth atomic coordinates in three-dimensional space. Given predicted coordinates $\hat{x}_i \in \mathbb{R}^3$ and ground-truth coordinates $x_i \in \mathbb{R}^3$ for $i = 1, \dots, N$, RMSD is defined as

$$
\mathrm{RMSD} \;=\; \sqrt{\frac{1}{N} \sum_{i=1}^{N}{||\hat{x}_i - x_i||^2_2}} 
$$

Prior to computation, predicted and reference structures are optimally aligned using a rigid-body superposition (Kabsch algorithm) to remove global translation and rotation. RMSD is sensitive to both local and global geometric errors and directly reflects reconstruction accuracy.

\paragraph{TM-score. \cite{zhang2004scoring}} 
TM-score evaluates the similarity of global structural topology while being relatively insensitive to structure length. It is defined as

$$
\mathrm{TM} \;=\; \frac{1}{L} \sum_{i=1}^{L} \frac{1}{1 + \left(\frac{d_i}{d_0(L)}\right)^2},
\quad
d_0(L) \;=\; 1.24\sqrt[3]{L-15}-1.8.
$$

where $L$ denotes the structure length, $d_i$ is the distance between the $i$-th pair of aligned residues, and 
$d_0(L)$ is a length-dependent normalization constant. TM-score takes values in $(0,\;1]$, with higher values indicating greater global structural similarity. Compared to RMSD, TM-score emphasizes correct overall folding rather than fine-grained local deviations.

\paragraph{Local Distance Difference Test (lDDT).} 
lDDT assesses the accuracy of local geometric relationships without requiring global structure alignment. The metric compares pairwise distances between atoms (or residues) in the predicted and reference structures.

For all atom pairs $(i,\;j)$ whose true distance $d_{ij}$ is below a fixed cutoff (typically 15 \AA), the absolute distance difference $|\hat{d}_{ij} - d_{ij}|$ is evaluated against multiple tolerance thresholds (e.g., 0.5 Å, 1 Å, 2 Å, and 4 Å). The lDDT score is computed as the average fraction of atom pairs satisfying these thresholds, yielding values in $[0,\;1]$.

lDDT is particularly sensitive to local structural correctness and complements global metrics such as RMSD and TM-score.

\subsection{Inverse Folding Metrics}
In the inverse folding task, the model generates RNA sequences conditioned on a given target structure. Performance is evaluated using sequence-level metrics.

\paragraph{Sequence Recovery. \cite{dauparas2022robust}}
Sequence recovery measures the fraction of positions at which the predicted sequence matches the ground-truth sequence exactly. It is defined as

$$\mathrm{Rec} \;=\; \frac{1}{L} \sum_{i=1}^{L} \mathbb{1}(\hat{s}_i = s_i).$$

where $s_i$ and $\hat{s}_i$ denote the ground-truth and predicted nucleotide identities at position $i$, respectively, and $\mathbb{1}(\cdot)$ is the indicator function. This metric reflects the model’s ability to reproduce native sequence preferences under structural constraints.

\paragraph{3-mer Diversity.}
Given $n$ candidate sequences designed for the same RNA backbone, we characterize sequence diversity using 3-mer frequency statistics. For each candidate sequence $s$, let $v_s \in \mathbb{R}^{64}$ denote the normalized frequency vector over all possible nucleotide 3-mers. We define the 3-mer diversity as

$$\mathrm{Div}_{3\mathrm{mer}} \;=\; 1 \;-\; \mathbb{E}_{s \neq t} [\rho(v_s, v_t)],$$

where $\rho(\cdot)$ denotes the Pearson correlation coefficient between two 3-mer frequency vectors. This metric quantifies sequence-level diversity by measuring the average dissimilarity of local substring usage across generated candidates. Higher values indicate greater diversity and reduced mode collapse.

\subsection{RNA–Ligand Binding Metrics.}
For the RNA–ligand binding task, we formulate the problem as binary classification, predicting whether a given RNA–ligand pair is binding or non-binding. Model performance is evaluated using the Area Under the Receiver Operating Characteristic Curve (AUROC).

AUROC is defined as the area under the receiver operating characteristic (ROC) curve, which plots the true positive rate (TPR) against the false positive rate (FPR) across all possible decision thresholds. Higher AUROC values indicate stronger binding specificity and reflect threshold-independent classification performance.

\section{Implementation Details}
This section provides details of the RiboSphere architecture, training pipeline, and inference procedure.

\subsection{Hyperparameters}
We train all models using AdamW with $\beta_1=0.9$, $\beta_2 = 0.95$. Unless otherwise specified, models are trained on 8 NVIDIA A6000 GPUs using data-parallel training. Gradient accumulation is applied to achieve a stable effective batch size across devices. Each micro-step processes a single RNA structure with randomly augmented rigid-body transformations. This design avoids padding or masking across variable-length sequences and simplifies batching during training.
\begin{table}[ht]
\centering
\caption{Training configuration hyperparameters.}
\label{tab:training_hyperparameters}
\begin{tabular}{l c}
\toprule
\textbf{Parameter} & \textbf{Value} \\
\midrule
Encoder layers & \textbf{2} /\ 4 /\ 6 /\ 8 \\
Decoder layers & 2 /\ 4 /\ 6 /\ \textbf{8} \\
Attention heads & 8 \\
Encoder channels & \textbf{256} /\ 512 \\
Decoder channels & 512 \\
Pair-bias channels & 64 \\
FSQ levels & (8, 6, 5) /\ (8, 5, 5, 5) /\ (7, 5, 5, 5, 5)\\
MLP factor & 4 \\
\midrule
Learning rate & $3 \times 10^{-4}$ \\
Epochs & 100 \\
Sliding window size & \textbf{8} /\ 16 /\ None \\
Gradient accumulation (micro-steps) & 8 \\
QK normalization & True /\ \textbf{False} \\
\bottomrule
\end{tabular}
\end{table}

\subsection{Data Preprocessing and Augmentation}
During training, all atomic coordinates are mean-centered to remove global translations, i.e.,

$$\tilde{x} \; = \; x - \frac{1}{N} \sum_{i=1}^{N} x_i ,$$

Since diffusion and flow-based generative processes cannot be defined in a translation-invariant manner over the full Euclidean space, we instead operate in the zero center-of-mass subspace. Concretely, if both the data sample $x$ and Gaussian noise $\epsilon \thicksim \mathcal{N}(0,\; \mathbf{I})$ satisfy $\sum_i x_i = \sum_i \epsilon_i = 0$, the noise interpolation used in flow matching,
$$x_t \; = \; tx + (1-t)\epsilon ,$$
remains well-defined and closed under linear combinations.

To improve generalization across global orientations, we augment RNA structures with random 3D rotations during training. For each RNA chain, multiple rotated copies are generated by sampling rotation matrices $R$ uniformly from SO(3) and applying them to the atomic coordinates as $x' = Rx$. This augmentation strategy preserves internal geometry while preventing the model from overfitting to specific global orientations.

Additional preprocessing steps include masking invalid or missing atoms, unit normalization, and optional selection of different atomic subsets, depending on the experimental setting.

\subsection{FSQ Details}
We implement the discrete bottleneck with finite scalar quantization (FSQ). Given an encoder output
$c_i \in \mathbb{R}^{d}$ for residue $i$, we first project it to the FSQ dimension and bound each
channel with a $\tanh$ nonlinearity. For the $k$-th FSQ channel with $L_k$ quantization levels, the
bounded scalar is rounded to the nearest integer:
\[
\hat{c}_{i,k} = \mathrm{round}\!\left(\left\lfloor \frac{L_k}{2} \right\rfloor
\tanh(W_k c_i)\right),
\]
where gradients through the rounding operation are propagated using the straight-through estimator.
The per-channel level configuration is denoted by
$\mathbf{L} = [L_1,\ldots,L_d]$, and defines an implicit Cartesian-product codebook
\[
\mathcal{C} = \prod_{k=1}^{d} \{0,\ldots,L_k-1\}, \qquad
|\mathcal{C}| = \prod_{k=1}^{d} L_k .
\]
Thus, each quantized vector $\hat{c}_i$ corresponds to one discrete structural token. We convert
$\hat{c}_i$ to a token index by mixed-radix enumeration:
\[
\mathrm{idx}(\hat{c}_i) =
\sum_{k=1}^{d} \tilde{c}_{i,k}
\prod_{m<k} L_m ,
\]
where $\tilde{c}_{i,k}$ denotes the shifted non-negative integer value of the $k$-th quantized
coordinate. This gives a bijection between FSQ code vectors and integer token IDs used by the
decoder and downstream modules.

In our experiments, different target vocabulary sizes are obtained by choosing different level
configurations. Specifically, we use $\mathbf{L}=(8,6,5)$ for a 240-token codebook,
$\mathbf{L}=(8,5,5,5)$ for a 1000-token codebook, and
$\mathbf{L}=(7,5,5,5,5)$ for a 4375-token codebook. Following the FSQ design, we keep each
per-channel level count at least five when possible, which provides a compact low-dimensional
quantizer while avoiding very coarse binary-style partitions.

\subsection{Flow Matching Inference Details}
At inference time, we generate RNA atomic coordinates by numerically integrating the learned conditional flow from an isotropic Gaussian prior to the data distribution. We adopt classifier-free guidance (CFG), score annealing, and noise annealing, following standard practices in diffusion-based generative modeling. Unless otherwise stated, all inference hyperparameters are inherited from ProteinFlow-style settings and are not extensively tuned, as the diffusion autoencoder framework allows the noise and score scales to be adjusted flexibly for different downstream tasks.

Concretely, given a sequence of discrete structural tokens, we initialize coordinates with zero-mean Gaussian noise and evolve them using an Euler discretization of the conditional flow matching dynamics. At each time step, we evaluate both the conditional vector field and an unconditional counterpart obtained by masking the codebook conditioning, and combine them using a classifier-free guidance weight. For intermediate time steps, stochasticity is injected through a Gaussian noise term whose variance is modulated by a time-dependent schedule, while near the terminal time the dynamics become fully deterministic.

\subsection{Inverse Folding Adapter Details}
The Adapter module is designed to map the latent discrete structural representations learned by RiboSphere to sequence generation tasks. It employs a dual-path architecture that maintains both scalar and vector features: the scalar path captures latent structural context, while the vector path explicitly models three-dimensional spatial directions to capture structural topology. Given latent node features $\hat{c}$, we first project them into scalar node features and vector-channel gates:
$$node_s \; = \; W_s \hat{c}, \quad v_{gate} \;=\; \text{tanh}(W_v\hat{c})$$
To enable selective modulation of geometric directions, a scalar-to-vector feedback is applied:

$$v_{weight} \; = \; v_{gate} + W_{s2v}(node_s),$$

which gates local directional vectors $v_{local}$ to form updated node vectors:

$$node_v \; = \; v_{weight} \odot \text{mean}_V(v_{local})$$

To capture pairwise geometric dependencies, a pairwise message-passing operator is used. Relative positional biases generate a pairwise relation matrix $S$, which is processed via a multi-layer perceptron and a causal mask $\mathcal{M}$:

$$v_{pair} \; = \; \sum_j (\sigma(S) \odot \mathcal{M})_{ij} v_{local}^j,$$

followed by aggregation:

$$node_v \leftarrow node_v + \text{mean}_V(v_{pair}).$$

Finally, vector-to-scalar feedback projects directional information back to the scalar path:

$$node_s \leftarrow node_s + W_{v2s}(node_v),$$

enhancing the scalar features’ sensitivity to spatial topology. This bidirectional interaction integrates latent structural context with explicit directional cues. The resulting geometry-aware node representations are passed to the autoregressive GVP decoder for nucleotide prediction.

\section{Additional Experimental Results}
\subsection{Extended Reconstruction Results}
Table \ref{tab: ex_results} presents extended reconstruction results for RiboSphere, complementing the main results reported in the text. These additional experiments provide further insights into the impact of different encoder/decoder configurations, embedding dimensions, and number of atoms on structural reconstruction quality and vector quantization usage.

\begin{table*}[t]
\centering
\caption{Extended Reconstruction Results of RiboSphere.} 
\label{tab: ex_results}
\resizebox{0.9\textwidth}{!}{
\begin{tabular}{l ccccccc cc}
\toprule
\multirow{2}{*}{\textbf{Method}} & \multirow{2}{*}{\textbf{\# Enc}} & \multirow{2}{*}{\textbf{\# Dec}} & \multirow{2}{*}{\textbf{Dim}} & \multirow{2}{*}{\textbf{\# Atoms}} & \multicolumn{3}{c}{\textbf{Structure}} & \multicolumn{2}{c}{\textbf{VQ}} \\
\cmidrule(lr){6-8} \cmidrule(lr){9-10}
& & & & & \textbf{RMSD} & \textbf{TM-score} & \textbf{lDDT} & \textbf{Codebook} & \textbf{\% Util.} \\
\midrule
E6-D6, D512, A1 & 6 & 6 & 512 & 1  & 3.03 & 0.63 & 0.70 & 240 & 100 \\
E6-D6, D512, A10 & 6 & 6 & 512 & 10 & 2.70 & 0.67 & 0.71 & 240 & 100 \\
E6-D6, D512, A11 & 6 & 6 & 512 & 11 & 2.89 & 0.65 & 0.71 & 240 & 100 \\
\midrule
E6-D6, D512, A1 & 6 & 6 & 512 & 1  & 1.88 & 0.73 & 0.75 & 1,000 & 95.0 \\
E6-D6, D512, A10 & 6 & 6 & 512 & 10 & 2.44 & 0.67 & 0.71 & 1,000 & 88.5 \\
E6-D6, D512, A11 & 6 & 6 & 512 & 11 & 2.71 & 0.68 & 0.72 & 1,000 & 98.9 \\
\midrule
E6-D6, D512, A1 & 6 & 6 & 512 & 1  & 2.16 & 0.73 & 0.76 & 4,375 & 39.2 \\
E6-D6, D512, A10 & 6 & 6 & 512 & 10 & 2.01 & 0.74 & 0.75 & 4,375 & 46.6 \\
E6-D6, D512, A11 & 6 & 6 & 512 & 11 & 2.34 & 0.68 & 0.71 & 4,375 & 35.6 \\
\midrule
E4-D6, D256, A11 & 4 & 6 & 256 & 11 & 2.43 & 0.67 & 0.72 & 240 & 100 \\
E4-D6, D512, A11 & 4 & 6 & 512 & 11 & 2.46 & 0.67 & 0.71 & 240 & 100 \\
\midrule
E4-D6, D256, A11 & 4 & 6 & 256 & 11 & 2.58 & 0.66 & 0.71 & 1,000 & 81.2 \\
E4-D6, D512, A11 & 4 & 6 & 512 & 11 & 3.10 & 0.64 & 0.70 & 1,000 & 80.5 \\
\midrule
E4-D6, D256, A11 & 4 & 6 & 256 & 11 & 2.70 & 0.71 & 0.74 & 4,375 & 39.8 \\
E4-D6, D512, A11 & 4 & 6 & 512 & 11 & 2.13 & 0.72 & 0.75 & 4,375 & 38.6 \\
\bottomrule
\end{tabular}%
}
\end{table*}

\subsection{Additional Inverse Folding Analysis}
To further characterize the behavior of our inverse folding model, we examined the trade-off between sequence recovery and diversity under different sampling temperatures. 

\begin{figure*}[t]
    \centering
    \includegraphics[width=0.40\linewidth]{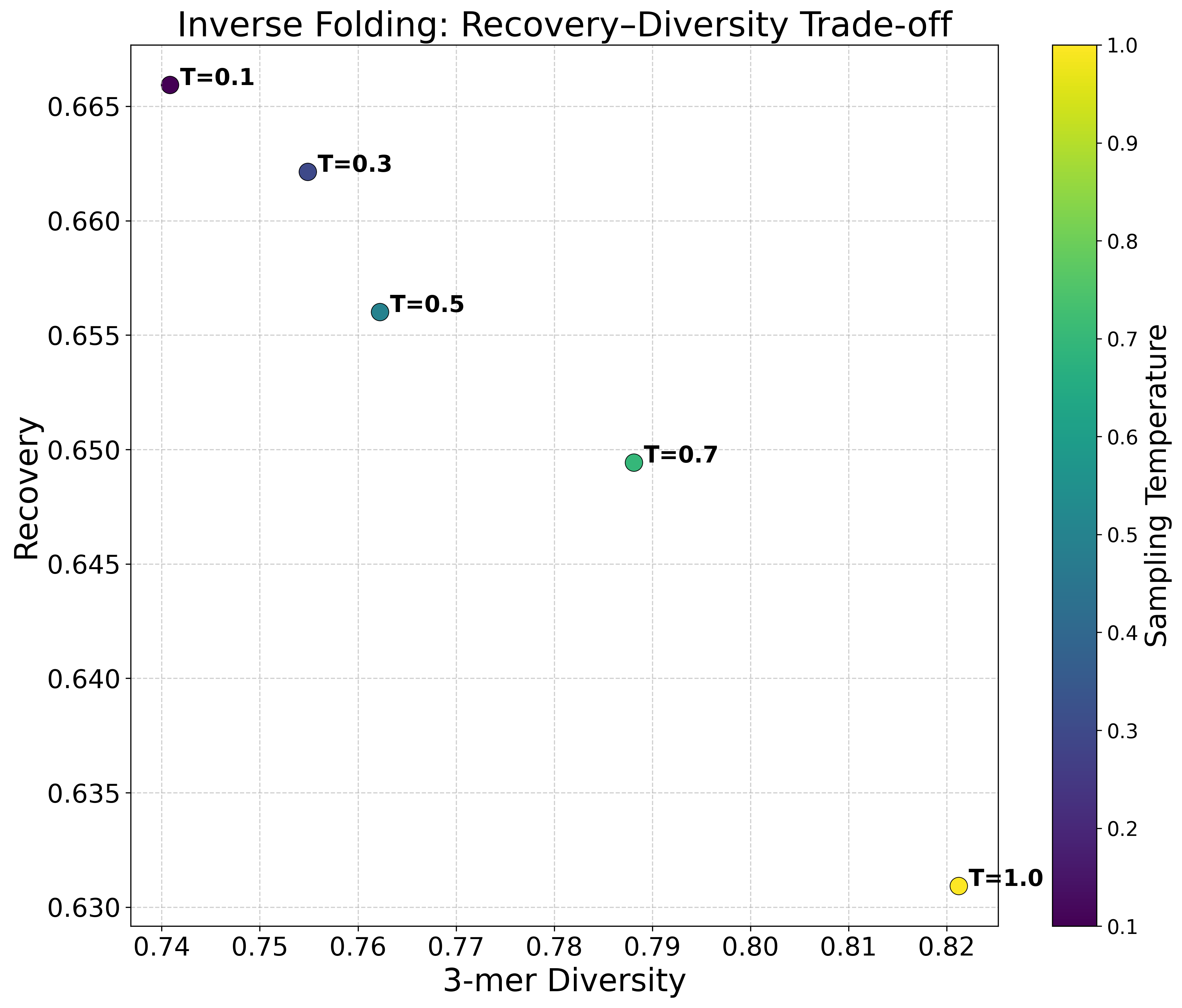} 
    \caption{Recovery–Diversity trade-off for the inverse folding model under different sampling temperatures.}
    \label{fig:recovery_diversity}
\end{figure*}

Figure~\ref{fig:recovery_diversity} shows the Recovery–Diversity trade-off curve obtained from sampling 16 sequences per structure at temperatures $T = \{0.1, 0.3, 0.5, 0.7, 1.0\}$. As expected, lower temperatures yield higher recovery rates, with $T=0.1$ achieving the highest mean recovery of $0.666$. Conversely, higher temperatures promote sequence diversity, with $T=1.0$ producing the largest 3-mer diversity of $0.821$. The observed trend demonstrates the inherent trade-off: improving sequence recovery generally reduces diversity, and vice versa. Intermediate temperatures around $T=0.5$–$0.7$ provide a balanced compromise between recovery and diversity.

Overall, this analysis highlights the flexibility of our model in generating multiple candidate sequences while controlling the trade-off via sampling temperature. This capability is particularly useful for downstream RNA design applications where both fidelity to the native structure and sequence variability are desirable.

\end{document}